\documentclass[letterpaper, 10 pt, conference]{ieeeconf}  

\IEEEoverridecommandlockouts                              

\usepackage{biblatex}
\usepackage{hyperref}
\usepackage{amsmath,amsthm,amssymb}

\usepackage{siunitx}
\usepackage{todonotes}
\usepackage{leftindex}
\usepackage{bm}
\usepackage{algorithmicx} 
\usepackage{algorithm}
\usepackage[noend]{algpseudocode}
\usepackage{caption}
\usepackage{subcaption}
\usepackage{graphicx}
\usepackage{gensymb} 
\usepackage{multirow}

\algrenewcommand\algorithmicrequire{\textbf{Input:}}
\algrenewcommand\algorithmicensure{\textbf{Output:}}

\usepackage[font=small,labelfont=bf]{caption}

\newtheorem{problem}{Problem}
\newtheorem{assumption}{Assumption}

\newtheorem{definition}{Definition}

\newtheorem{proposition}{Proposition}

\title{\LARGE \bf
One-Shot Transfer of Long-Horizon Extrinsic Manipulation \\ Through Contact Retargeting
}

\author{Albert Wu$^{1}$, Ruocheng Wang$^{1}$, Sirui Chen$^{1}$, Clemens Eppner$^{2}$, and C. Karen Liu$^{1}$
\thanks{$^{1}$Computer Science Department, Stanford University, Stanford, CA 94305, USA
        {\tt\small \{amhwu, rcwang, ericcsr ,karenliu\}@cs.stanford.edu}}%
\thanks{$^{2}$NVIDIA, Seattle, WA 98105, USA
        {\tt\small ceppner@nvidia.com}}%
}

\begin{document}

\maketitle
\thispagestyle{empty}
\pagestyle{empty}

\begin{abstract}
Extrinsic manipulation, the use of environment contacts to achieve manipulation objectives, enables strategies that are otherwise impossible with a parallel jaw gripper. However, orchestrating a long-horizon sequence of contact interactions between the robot, object, and environment is notoriously challenging due to the scene diversity, large action space, and difficult contact dynamics. We observe that most extrinsic manipulation are combinations of short-horizon primitives, each of which depend strongly on initializing from a desirable contact configuration to succeed. Therefore, we propose to generalize one extrinsic manipulation trajectory to diverse objects and environments by retargeting contact requirements. We prepare a single library of robust short-horizon, goal-conditioned primitive policies, and design a framework to compose state constraints stemming from contacts specifications of each primitive. Given a test scene and a single demo prescribing the primitive sequence, our method enforces the state constraints on the test scene and find intermediate goal states using inverse kinematics. The goals are then tracked by the primitive policies. Using a 7+1 DoF robotic arm-gripper system, we achieved an overall success rate of 80.5\% on hardware over 4 long-horizon extrinsic manipulation tasks, each with up to 4 primitives. Our experiments cover 10 objects and 6 environment configurations. We further show empirically that our method admits a wide range of demonstrations, and that contact retargeting is indeed the key to successfully combining primitives for long-horizon extrinsic manipulation. Code and additional details are available at \url{stanford-tml.github.io/extrinsic-manipulation}.
\end{abstract}


\section{Introduction}
Extrinsic manipulation describes the usage of environment contact to aid manipulation~\cite{raessa2019visually} and is an emerging field in robotic manipulation research. Leveraging environment contacts allows simple parallel jaw grippers to achieve complex tasks that are otherwise impossible. For instance, objects in ungraspable initial poses can be picked up by first executing a sequence of pregrasp motions involving pushing, pivoting, and pulling~\cite{chen2023synthesizing, zhou2023learning, yang2023learning}.

Achieving extrinsic manipulation requires a holistic orchestration of contacts interactions between the robot, object, and environment. In particular, the robot must be able to address the diverse object and environment geometries present in the real world. Earlier works attempt to perform control synthesis by modeling contact dynamics explicitly~\cite{costanzo2020grasp, doshi2022manipulation, aceituno2020global}. However, due to the inherent difficulty with modeling contact dynamics, these works are restricted to manipulating known objects with simple geometries. Recent literature on extrinsic manipulations have sought to produce reinforcement learning(RL)-based extrinsic manipulation policies that generalizes to novel objects ~\cite{zhou2023learning, kim2023pre, zhou2023hacman, cho2024corn,yang2023learning}, but none has successfully generalized to novel environments to our best knowledge. 
\begin{figure}[!tb]
    \centering
    \includegraphics[width=0.48\textwidth]{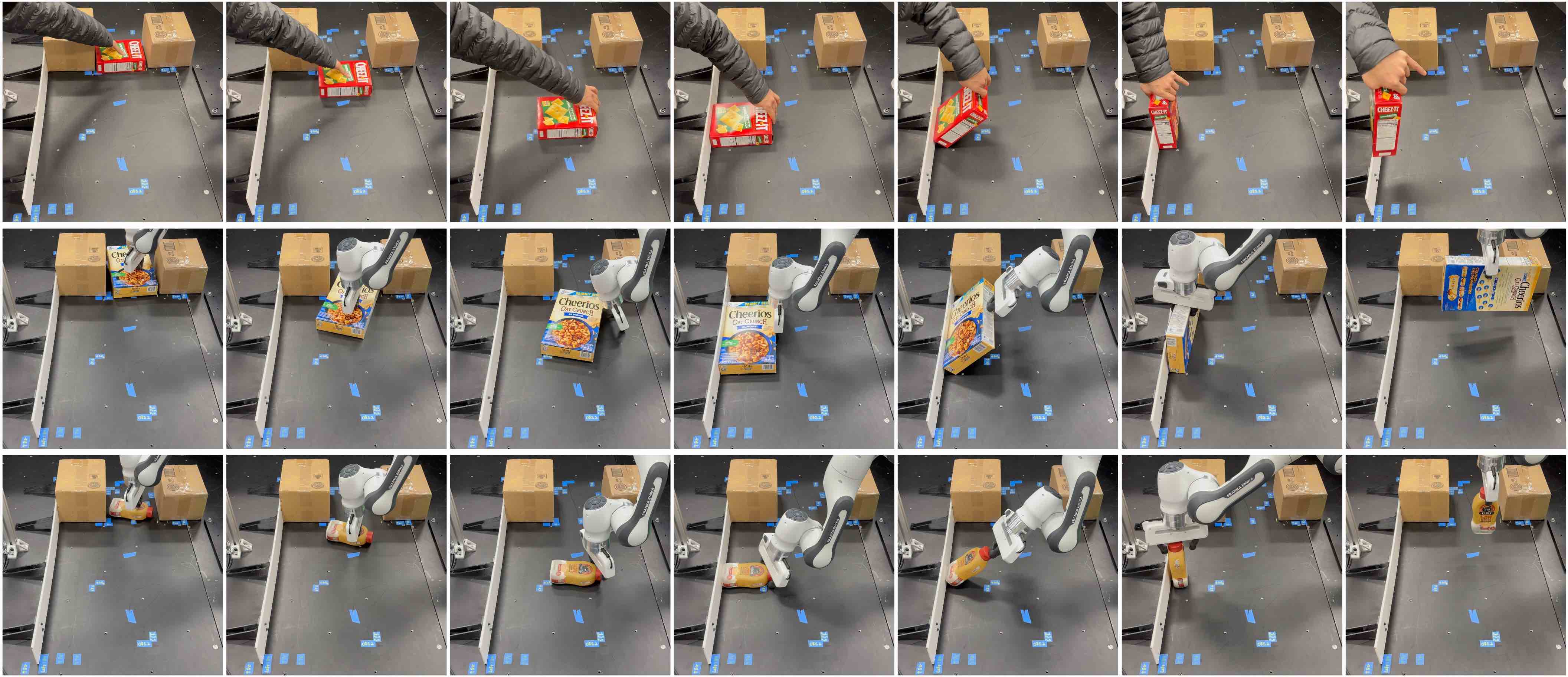}
    \caption{\textbf{Retargeting the \textit{object retrieval} task} from a human demo (top) to \textit{oat} (middle) and \textit{flapjack} (bottom). This 4-primitive task involves \textit{pulling} the object from between the obstacles, \textit{pushing} it to the wall, \textit{pivoting} against the wall to expose a graspable edge, and finally \textit{grasping} the object. Each row shows a trajectory in temporal order from left to right. Please refer to our supplementary video and website for animations.}
    \label{fig:highlight_retrieval}
\end{figure}

Furthermore, many applications in manipulation, such as occluded grasping~\cite{zhou2023learning}, are long-horizon in nature and require multiple contact switches. To overcome the difficulty in producing long-horizon plans, some works in literature seek to leverage demonstration~\cite{florence2022implicit, chi2023diffusion,mandlekar2020learning}. However, to achieve achieve contact-rich manipulation, hundreds of demonstrations for a single task may be necessary~\cite{florence2022implicit, chi2023diffusion}. 

Other works seeking to achieve long-horizon manipulation leverage a hierarchical structure to abstract away repetitive, low-level motion ``primitives''~\cite{nasiriany2022augmenting, gao2024prime, dalal2021accelerating, chen2023predicting}. While we observe that most long-horizon extrinsic manipulation is nothing more than a combination of short-horizon, fixed-contact configuration primitives, for instance ``push'' and ``pivot'', extrinsic manipulation primitives are significantly more challenging to abstract than simple primitives such as pick, place, or end effector movement~\cite{nasiriany2022augmenting, gao2024prime}. Extrinsic manipulation primitives leverage sophisticated contact interactions, for instance ``pivot'' requires an object to be in contact with an environment obstacle. This imposes scene-dependent contact preconditions and exacerbates the sim-to-real gap. Nevertheless, once the contact precondition is satisfied, an extrinsic manipulation primitive is much more likely to succeed.

We therefore propose to solve two subproblems instead of the full long-horizon extrinsic manipulation problem: obtaining primitives for a given contact configuration that are robust under object and environment variations, and initializing each primitive in the desired contact configuration. Moreover, we observe that the sequence of primitives is typically fixed for a given manipulation objective. For instance, ``occluded grasping by pushing object against an obstacle and pivoting to expose graspable edge'' always has a sequence of ``push-pivot-grasp.'' Hence, one may circumvent the combinatorially complex contact sequence planning by collecting a task demonstration that fixates the contact sequence.

In this paper, we describe a method to generalize long-horizon extrinsic manipulation plans from a single demonstration. Our contributions are:\\
$\bullet\;\,$\textbf{Contact retargeting framework} to respect contact constraints while chaining extrinsic manipulation primitives. We formalize a process to merge semantic contact specifications from adjacent primitives, and leverage inverse kinematics (IK) to find states achieving such requirements.\\ 
$\bullet\;\,$\textbf{One-shot transfer pipeline of extrinsic manipulation task demo} to diverse object and environment. We prepare one library of 4 extrinsic manipulation primitives robust to scene variations. By leveraging contact retargeting, our pipeline merely takes a single task demo of any primitive combination to achieve the same task in a distinct scene.
\\ 
$\bullet\;\,$\textbf{Extensive hardware validation over 4 long-horizon tasks} covering 10 objects and 4 environments using a wide range of demos. Our method achieved an overall success rate of $80.5\%$ and outperformed~\cite{zhou2023learning} on occluded grasping. Ablation shows contact retargeting is indeed the key to successfully chaining extrinsic manipulation primitives.

\section{Related work}

\subsection{Manipulation using environment contacts}
Many works in literature have explored applications that require environment contacts, such as in-hand repositioning with a simple gripper~\cite{dafle2014extrinsic, stepputtis2018extrinsic}, grasping from an initially ungraspable pose~\cite{costanzo2020grasp,zhou2023learning, yang2023learning,sun2020learning,eppner2015exploitation,pollayil2021planning}, and realignment for industrial assembly~\cite{raessa2019visually}. Typically, the manipulated object is pushed along or slid against a flat surface in the environment, such as a tabletop or a wall~\cite{costanzo2020grasp,zhou2023learning,kim2023pre,zhou2023hacman,doshi2022manipulation,cho2024corn,yang2023learning,hou2019robust,eppner2015exploitation}. In some cases, gravity is also leveraged for in-hand reorientation~\cite{dafle2014extrinsic, stepputtis2018extrinsic}.

To produce these motions, earlier works often synthesize control policies through hand-designed strategies~\cite{raessa2019visually,costanzo2020grasp,eppner2015exploitation} or physical models~\cite{dafle2014extrinsic, costanzo2020grasp, doshi2022manipulation, aceituno2020global,hou2019robust, cheng2022contact,pollayil2021planning}. However, the inherent difficulty of modeling contact-rich motion restricts these works to manipulating known objects with simple geometries. Other papers recent papers instead uses RL to obtain a feedback policy~\cite{zhou2023learning, kim2023pre, zhou2023hacman, cho2024corn,yang2023learning,sun2020learning}, which can be significantly more robust and generalize to geometries never seen during training~\cite{zhou2023learning,zhou2023hacman,cho2024corn,sun2020learning}. Attention to variable environments, meanwhile, is much more scarce. Among the papers reviewed in this section, only ~\cite{cheng2022contact, sun2020learning} consider environment variability. The discussion in ~\cite{cheng2022contact} is limited to simulation experiments, and ~\cite{sun2020learning} uses a fingerless, ball-shaped end effector with no grasping capability. 

\subsection{Composing primitives for long-horizon manipulation}
Divide-and-conquer has been a popular way to tackle long-horizon manipulation in literature. The learning-based robotics community typically leverage the framework of Parameterized Action MDP~\cite{masson2016reinforcement} to break down tasks into shorter horizon primitives. The primitives can be manually specified~\cite{dalal2021accelerating} or learned~\cite{nasiriany2022augmenting}. A high-level policy, which may be based on RL~\cite{dalal2021accelerating,nasiriany2022augmenting}, Graph Neural Network~\cite{chen2023predicting}, or imitation learning~\cite{gao2024prime} is then used to determine the sequence and parameters of the primitives. 

The main challenge with primitive-based approaches is ensuring the composability of the primitives. Many advanced primitives cannot succeed from arbitrary initial states and experience a large sim-to-real gap. ~\cite{dalal2021accelerating} has no hardware experiments.~\cite{nasiriany2022augmenting, gao2024prime} only have hardware experiments on pick-and-place tasks and end effector movement. ~\cite{chen2023predicting} can only perform a single stowing task. \cite{chen2023sequential} uses dexterous manipulation primitives that must be fine-tuned for a specific execution sequence. Composing complex primitives arbitrarily is still a missing ability in literature.

We note that primitive sequence planning, which is considered in~\cite{dalal2021accelerating, nasiriany2022augmenting, chen2023predicting}, is beyond the scope of this work. As such, this section excludes the task and motion planning (TAMP) literature, which has an emphasis on planning such sequences. The interested reader is referred to~\cite{garrett2021integrated} for a thorough review on TAMP. Our contribution focuses on assuring the feasibility of the primitive sequence; choice of the discrete sequence is obtained directly from the single demonstration.

\subsection{Demonstration for manipulation}
Leveraging demonstration for robotic manipulation has gained significant traction in recent years. 
The robot learning community has explored many approaches, including variations of behavior cloning (e.g. ~\cite{florence2022implicit, wen2022you, chi2023diffusion,mandlekar2020learning}) and fine-tuning a pretrained RL agent (e.g. ~\cite{sontakke2024roboclip, singh2020parrot}). We refer the readers to~\cite{ravichandar2020recent} for a thorough review.

While these methods have achieved many sophisticated manipulation tasks, the scalability and generalizability of the policies produced by these methods is unclear. Despite novel systems such as ~\cite{fu2024mobile, chi2024universal} lowering the barrier to collect demonstrations, the sheer amount of data needed means training a policy is still costly. Every task in ~\cite{chi2023diffusion, florence2022implicit, chi2024universal} requires between 50 to over 500 demonstrations. Some works such as~\cite{wen2022you, johns2021coarse, sontakke2024roboclip} emphasize the use of single demonstrations, but the tasks achieved by these approaches are restricted to end-effector pose movement with a firmly grasped object. Achieving complex manipulation tasks from few demonstrations remains an open problem to the best of our knowledge.






\section{Notation and problem definition}
We consider the quasi-static extrinsic manipulation of an object with geometry $\mathcal{O}\in\bm{\mathcal{O}}$ in environment $\mathcal{E}\in\bm{\mathcal{E}}$ using a 7-degree-of-freedom (DoF) robotic arm with a 1DoF parallel jaw gripper. The robot system has state $\bm{q} \in\mathcal{Q}\subseteq\mathbb{R}^{8}$, and the pose of the object is $\bm{x}\in \mathcal{X} \subseteq SE(3)$. We adopt the shorthand $\bm{s}_t \triangleq (\bm{x}_{t}, \bm{q}_{t})\in\mathcal{S}\triangleq\mathcal{X}\times\mathcal{Q}$ to denote the state of the system. Denote the control action as $\bm{u}\in\mathcal{U}\subset\mathbb{R}^8$. The system is governed by the environment and object dependent, discretized dynamics:
$\bm{s}_{t+1} = \leftindex^{\mathcal{E}, \mathcal{O}}f(\bm{s}_t, \bm{u}_t)$.
A $T+1$ step trajectory can thus be described as $
\left\{^{\mathcal{E},\mathcal{O}}\bm{s}_t\right\}_{t=0,1,\dots,T}$.

We first make assumptions that are commonly satisfied in manipulation scenarios.
\begin{assumption}
\label{assump:fixed_environment}
$\mathcal{E}$ is fixed throughout the task.
\end{assumption}
\begin{assumption}
\label{assump:fully_actuated}
The robotic manipulator is fully actuated. A motion planner is available such that the robot can reach anywhere within the workspace without collisions and does not obstruct any object-environment contacts.
\end{assumption}

To concretely describe contact configurations in the system, we adopt Definitions~\ref{def:semantic_contact} and ~\ref{def:contact_states}.

\begin{definition}[Contact configuration]
    A contact configuration describes a \textbf{minimal set} of semantic contact requirements, for instance ``robot finger in contact with the top of an object; object in contact with the obstacle.'' We denote this as $\sigma\in\Sigma$. We use $\sigma^{\bm{x}}$ and $\sigma^{\bm{q}}$ to denote the subsets of $\sigma$ concerning semantic environment-object and robot-object contact configurations. $\sigma^{\bm{x}}\cup \sigma^{\bm{q}} = \sigma$. Under Assumption~\ref{assump:fully_actuated}, robot-environment contact is not considered. While our flexible framework allows each manipulation primitive to define its own contact requirements, we define a small set of general contact configurations used to implement all the manipulation primitives in this paper (Section~\ref{subsec:implement_sigma}).
    \label{def:semantic_contact}
\end{definition}
\begin{definition}[Contact configuration in $\mathcal{E},\mathcal{O}$]
    For a specific $\mathcal{E},\mathcal{O}$, ${}^{\mathcal{E},\mathcal{O}}\sigma\subset\mathcal{S}$ concretely defines a set of robot and object states that satisfies such semantic contact requirements. By assumption~\ref{assump:fully_actuated}, environment-object contact is independent of the robot configuration, thus ${}^{\mathcal{E},\mathcal{O}}\sigma^{\bm{x}}\subset \mathcal{X}$ defines a set of object states that satisfy $\sigma^{\bm{x}}$.    
    On the other hand, robot-object contact is dependent on both the object state and robot state. Thus we denote  ${}^{\mathcal{E},\mathcal{O}}\sigma^{\bm{q}\mid\bm{x}}\subset\mathcal{Q}$ as the set of robot states that satisfy $\sigma^{\bm{q}}$ given a particular $\bm{x}$.
    \label{def:contact_states}
\end{definition}





To allow abstracting the robot-object interaction away and admitting demonstrations performed by arbitrary end effectors, we define the following:
\begin{definition}[Freestanding object states]
For a given $\mathcal{E,O}$, the ``freestanding object states'' are objects states that stay unchanged indefinitely unless there is robot-object contact. We denote this as ${}^{\mathcal{E,O}}\mathcal{X}_{s}$.
\end{definition}

\begin{problem}[Retargeting extrinsic manipulation]
    Consider an extrinsic manipulation trajectory $\left\{^{\mathcal{\bar{E}},\mathcal{\bar{O}}}\bm{\bar{s}}_t\right\}_{t=0,1,\dots,T}$ collected in environment $\mathcal{\bar{E}}$ on object $\mathcal{\bar{O}}$ under
        $\bar{\bm{s}}_{t+1} = {}^{\bar{\mathcal{E}}, \bar{\mathcal{O}}}f(\bar{\bm{s}}_t, \bar{\bm{u}}_t).$
    The trajectory satisfies a manipulation objective $\mathcal{G} \subseteq \mathcal{X}, \; \bar{\bm{x}}_T\in\mathcal{G}$. Additionally, the trajectory progresses through $N$ contact configurations $\sigma_1,\dots, \sigma_N$, and has $n-1$ contact switches at $0<t_1<\dots<t_{N-1}<T$ at freestanding states, i.e.
    \begin{align}
    \bar{\bm{s}}_{t}&\in
    \leftindex^{\mathcal{\bar{E}},\mathcal{\bar{O}}}\sigma_{i},\; \forall t_i < t < t_{i+1}, \label{eqn:fixed_contact}
    \\
    \bar{\bm{x}}_{t_i} &\in \leftindex^{\mathcal{\bar{E}},\mathcal{\bar{O}}}\sigma_{i}^{\bar{\bm{x}}} \cap \leftindex^{\mathcal{\bar{E}},\mathcal{\bar{O}}}\sigma^{\bar{\bm{x}}}_{i+1}\cap {}^{\bar{\mathcal{E}}, \bar{\mathcal{O}}}\mathcal{X}_{s} \label{eqn:contact_switch}
    \end{align}


Given a test environment $\mathcal{E}$ and a test object $\mathcal{O}$, find a policy $\bm{\pi}(\cdot)$
such that, under a strictly-increasing time-remapping function $\eta:\mathbb{R}\mapsto\mathbb{R},\;\tau = \eta(t),\; \eta(0)\equiv 0,\; d\eta/dt>0$ 
, $\mathcal{O}$ achieves $\mathcal{G}$ in $\mathcal{E}$
\begin{align}
    \bm{s}_{\tau+1} &= \leftindex^{\mathcal{E}, \mathcal{O}}f\left(\bm{s}_\tau, \bm{\pi}(\bm{s}_\tau)\right), \;
    \bm{x}_{\eta(T)}\in\mathcal{G},
\end{align}
using the same contact sequence
\begin{align}
    \bm{s}_{\eta(t)}&\in
    \leftindex^{\mathcal{E},\mathcal{O}}\sigma_{i},\; \forall t_i < t < t_{i+1}\\
\bm{x}_{\eta(t_i)} &\in \leftindex^{\mathcal{E},\mathcal{O}}\sigma_{i}^{\bm{x}} \cap \leftindex^{\mathcal{E},\mathcal{O}}\sigma^{\bm{x}}_{i+1}\cap {}^{\mathcal{E,O}}\mathcal{X}_s.
\end{align}
\label{prob:ext_manip_retargeting}
\end{problem}

Lastly, we make Assumption~\ref{assump:fs_contact_solution} to ensure the problem can be solved with the same manipulation strategy as the demo.
\begin{assumption}
    Problem~\ref{prob:ext_manip_retargeting} can be solved using contact switches at freestanding states, i.e.
\begin{equation*}
{}^{\mathcal{E}, \mathcal{O}}\mathcal{X}_{s} \cap \leftindex^{\mathcal{E},\mathcal{O}}\sigma_{i}^{\bm{x}} \cap \leftindex^{\mathcal{E},\mathcal{O}}\sigma^{\bm{x}}_{i+1} \neq \emptyset, \forall i=1,\dots,N-1.
\end{equation*}
\label{assump:fs_contact_solution}
\end{assumption}

\section{Generalizing extrinsic manipulation demos with contact retargeting}
\begin{figure}[h]
    \centering
    \includegraphics[width=0.48\textwidth]{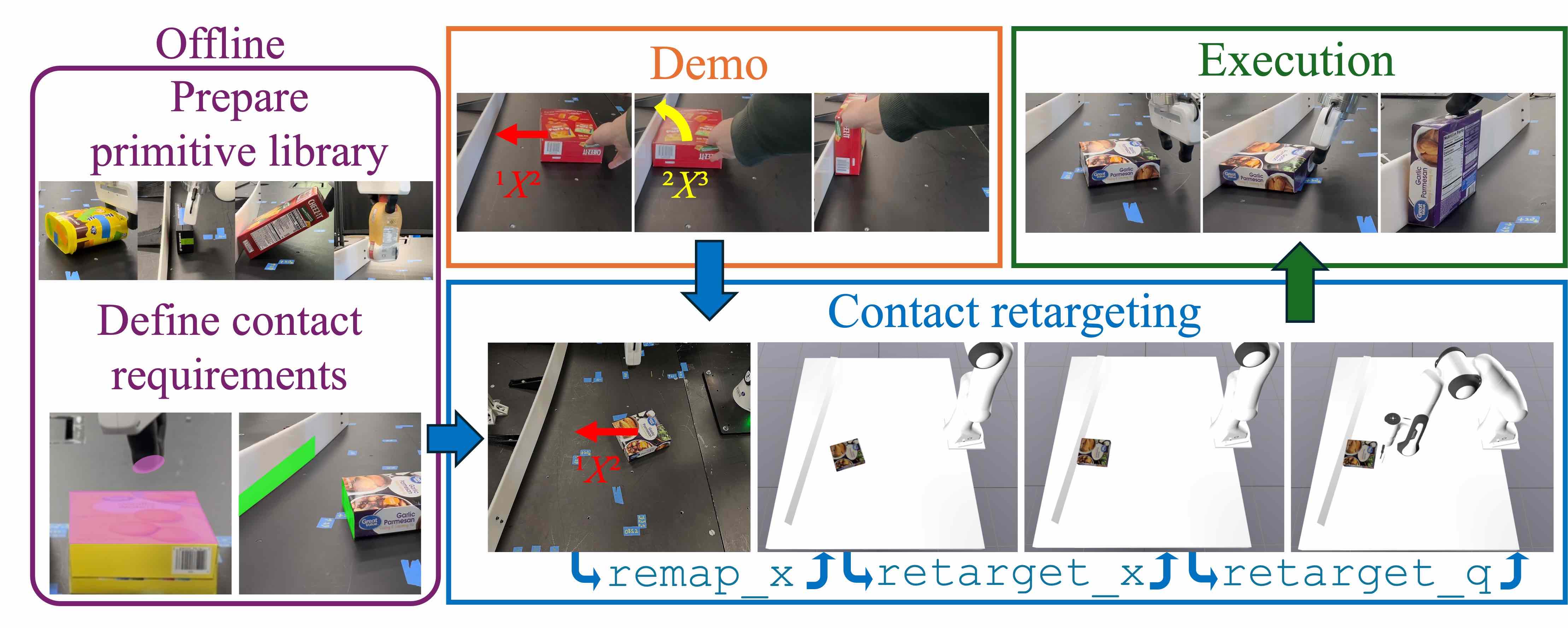}
    \caption{\textbf{Approach overview}. We prepare a primitive library and define each primitive's contact requirements online. Given a demo task trajectory and a test scene, we retarget the demo to the test scene by enforcing contact requirements. The demo's primitive sequence is then used to perform the task in the test scene.}
    \label{fig:approach}
\end{figure}

Our approach is grounded in the following observation:
\begin{enumerate}
    \item Long-horizon extrinsic manipulation can be decomposed into a sequence of primitives based on contact switches. \label{obs:primitive_decomp}
    \item A small set of primitives is sufficient to capture most extrinsic manipulation. \label{obs:small_primitive_set}
    \item The success of each primitive is highly dependent on satisfying the desired $\sigma$. \label{obs:contact_importance}
    \item Under Assumption~\ref{assump:fully_actuated} and~\ref{assump:fs_contact_solution}, satisfying $\sigma$ only entails selecting an environment-dependent $\bm{x}$ \label{obs:obj_state_contact}. 
\end{enumerate}
Observations~\ref{obs:primitive_decomp} and~\ref{obs:small_primitive_set} motivates building a primitive library to handle different scenarios. Observations~\ref{obs:contact_importance} and~\ref{obs:obj_state_contact} suggests that the key to generalizing extrinsic manipulation from $\bar{\mathcal{E}}, \bar{\mathcal{O}}$ to $\mathcal{E, O}$ is to find object states that all $\sigma$s are maintained. We thus break down Problem~\ref{prob:ext_manip_retargeting} into the following steps:
\begin{enumerate}
    \item Prepare a library of short-horizon primitives for any $\mathcal{E, O}$ (Sec.~\ref{subsec:primitive_library},~\ref{subsec:implement_sigma})
    \item Identify the primitive sequence in the demonstration (Sec.~\ref{subsec:demon_collection})
    \item Remap the object states from $\bar{\mathcal{E}}, \bar{\mathcal{O}}$ to $\mathcal{E, O}$ (Sec.~\ref{subsec:contact_retageting})
    \item Combine the sequence of remapped primitive sequence to achieve $\mathcal{G}$ (Sec.~\ref{subsec:policy_comp})
\end{enumerate}

We summarize our approach in Fig.~\ref{fig:approach}. The rest of this section describes each of the steps in detail. In Section~\ref{sec:experiments}, we show empirically that this is a much more tractable formulation for generalizing long-horizon extrinsic manipulation.

\subsection{Primitive library}
\label{subsec:primitive_library}
We build a primitive library $\Pi$ to handle short extrinsic manipulation tasks that start from a single contact configuration $\sigma_i$. In particular, we seek to develop a goal-conditioned policy $\bm{\pi}_{\sigma_i}$ where, given any $\mathcal{E}\in\bm{\mathcal{E}}, \mathcal{O}\in\bm{\mathcal{O}}$, initial state $\bm{s}_0=(\bm{x}_0,\bm{q}_0)\in {}^{\mathcal{E}, \mathcal{O}}\sigma_i$, and intermediate manipulation objective $\mathcal{G}_i \subseteq{}^{\mathcal{E}, \mathcal{O}}\sigma^{\bm{x}}_{i}\subset \mathcal{X}$, the state evolution
\begin{equation}
    \bm{s}_{t+1} = \leftindex^{\mathcal{E}, \mathcal{O}}f\left(\bm{s}_t, \bm{\pi}_{\sigma_i}(\bm{s}_t;\mathcal{E}, \mathcal{O}, \mathcal{G}_i)\right),
\end{equation}
eventually leads to a state in $\mathcal{G}_i$. Table~\ref{tab:primitives} summarizes our 4 primitives, \textit{push, pivot, pull, grasp}. The primitives may be obtained from any technique, including model-based and learning-based, and leverage different control method. Our primitives use joint impedance control and operational space control (OSC). The restriction of initializing in $\sigma_i$ and the short task horizon makes producing $\pi_{\sigma_i}$ for arbitrary $\mathcal{E}, \mathcal{O}$ significantly easier. Our primitive library is scalable: to add an additional primitive, one only need to provide $\pi_{\sigma_i}$ and implement $\sigma_i$ (Section~\ref{subsec:implement_sigma}).

We note that a robustly designed $\pi_{\sigma_i}$ may relax the requirements specified by $\sigma_i$. Enforcing contact configurations often requires real-time feedback control. To increase hardware performance, $\pi_{\sigma_i}$ may be designed such that states in the vicinity of ${}^{\mathcal{E},\mathcal{O}}\sigma$ can still successfully execute. Our pivot and pull primitives leverage compliance and feedback control to ensure the robot-object and environment-object contacts are maintained. We also discovered the pushing policy performs better without specifying the robot-object contact; the RL policy is able to choose the contact implicitly.



\begin{table*}[h]
\vspace{5pt}
\caption{Extrinsic manipulation primitives.}
\label{tab:primitives}
\begin{center}
\begin{tabular}{|c|c|c|c|c|}
\hline
 & Push & Pull & Pivot & Grasp\\
\hline
Description & Push from side & Pull from top & Pivot against environment & Top-down grasp \\
\hline
Type & RL policy & Designed & Designed & Designed  \\
\hline
Control type & Joint impedance & OSC & OSC & OSC \\
\hline
State feedback & $\bm{x,q}$ & Open loop & $\bm{q}$ & $\bm{x}$\\
\hline
Environment contact requirements $\sigma^{\bm{x}}$ & Ground & Ground & Ground, wall& Ground \\ 
\hline
Robot contact requirements $\sigma^{\bm{q}}$& $\emptyset$ (decided by $\bm{\pi}$) & Top & Antipodal & Grasp\\
\hline
\end{tabular}
\end{center}
\end{table*}

\subsection{Implementing $\sigma$}
\label{subsec:implement_sigma}
We implement $\sigma$ as state constraints in the inverse kinematics (IK) problem detailed in Section~\ref{subsec:contact_retageting}. States satisfying the constraints here are states in the set ${}^{\mathcal{E,O}}\sigma$, though the set is not computed explicitly in practice. The following environment-object constraint types are implemented for $\sigma^{\bm{x}}$.
\begin{enumerate}
    \item {\textbf{Ground}}: the object is in contact with the ground.
    \item {\textbf{Wall}}: a lower edge of the object is in contact and orthogonal to the wall normal. This is implemented using the bounding box of the object. Of the 4 vertices that are closest to the wall, the 2 lowest ones must be on the wall, and the distance between the wall and the object is 0.
\end{enumerate}
The following robot-object constraint types are defined for $\sigma^{\bm{q}\mid\bm{x}}$. As $\bm{x}$ is given, constraints specified relative to the world frame and environment are well defined.
\begin{enumerate}
    \item {\textbf{Top}}: the gripper fingertips are approximately in contact with the ``top'' of the object. This is implemented by finding the 4 vertices with the highest world $z$ coordinate of the object's bounding box, then finding the geometric center of the 4 points. The robot-object contact point is found by moving the end effector from geometric center along world $-z$ until contact established.
    \item {\textbf{Antipodal}}: the gripper fingertips are in contact with the object on the opposite side of the wall-object contact. This is implemented as an intersection of two constraints: the distances between the object and the fingertips are zero; the fingertips are within a cone centered at the object's geometric center, has the wall's normal as its axis, and has a half-angle of $\pi/6$.
    \item {\textbf{Grasp}}: the end effector is placed such that closing the gripper results in a top-down object grasp.
\end{enumerate}
The usage of these constraints in our extrinsic manipulation primitives is summarized in Table~\ref{tab:primitives}.

\subsection{Contact retargeting}
\label{subsec:contact_retageting}
We propose a \textit{contact retargeting} procedure to ensure each primitive is started from a state in its contact configuration regardless of $\mathcal{E,O}$. First, we observed from Assumption~\ref{assump:fs_contact_solution} that the robot $\bm{q}$ can change arbitrarily during each primitive switch:
\begin{proposition}[Arbitrary robot-object contact switch at a freestanding state]
    $\forall \bm{x}\in {}^{\mathcal{E}, \mathcal{O}}\mathcal{X}_{s} \cap \leftindex^{\mathcal{E},\mathcal{O}}\sigma_{i}^{\bm{x}} \cap \leftindex^{\mathcal{E},\mathcal{O}}\sigma^{\bm{x}}_{i+1}$, there exists $\bm{q}_1 \in \leftindex^{\mathcal{E},\mathcal{O}}\sigma_{i}^{\bm{q}\mid\bm{x}}$ and $\bm{q}_2 \in \leftindex^{\mathcal{E},\mathcal{O}}\sigma_{i+1}^{\bm{q}\mid\bm{x}}$ such that $(\bm{x}, \bm{q}_1)\in{}^{\mathcal{E,O}}\sigma_{i}$ and $(\bm{x},\bm{q}_{2})\in{}^{\mathcal{E,O}}\sigma_{i+1}$.
\label{prop:robot_object_contact_switch}
\end{proposition}

By Proposition~\ref{prop:robot_object_contact_switch}, in order to start the next primitive $\pi_{\sigma_{i+1}}$ from a state in ${}^{\mathcal{E}, \mathcal{O}}\sigma_{i+1}$, we only need the previous primitive to end at an object state in ${}^{\mathcal{E}, \mathcal{O}}\sigma^{\bm{x}}_{i+1}$. The robot can be moved subsequently to satisfy ${}^{\mathcal{E}, \mathcal{O}}\sigma^{\bm{q}}_{i+1}$. This leads to a two-stage contact retargeting subroutine: \texttt{retarget\_x} and \texttt{retarget\_q}. Both are implemented as IK problems using Drake~\cite{drake}. Additionally, we map the demo object states to the test object using \texttt{remap\_x} to serve as the initial guess for \texttt{retarget\_x}.

\texttt{remap\_x} extracts the relative transforms $\left\{{}^{i}X{}^{i+1}\right\}$ in the demo object states $\left\{\bar{\bm{x}}_{t_i}\right\}$ and apply them to the test object's initial state $\bm{x}_0$ to obtain a sequence of initial guess object states $\left\{\tilde{\bm{x}}_{i}\right\}$, i.e.
\begin{align}
    \bar{\bm{x}}_{t_{i+1}}  = {}^{i}X{}^{i+1}(\bar{\bm{x}}_{t_i}),\;\;
    \tilde{\bm{x}}_{i+1}  \triangleq {}^{i}X{}^{i+1}(\bm{x}_i).
\end{align}

\texttt{retarget\_x} finds $\mathcal{G}_i \subseteq {}^{\mathcal{E},\mathcal{O}}\sigma_{i}^{\bm{x}} \cap {}^{\mathcal{E},\mathcal{O}}\sigma^{\bm{x}}_{i+1}$ given  $\sigma_{i}^{\bm{x}}, \sigma^{\bm{x}}_{i+1}, \mathcal{E,O}, \tilde{\bm{x}}_{i+1}$ by solving the following IK problem:
\begin{align}
    \min_{\bm{x}}\; d(\bm{x}, \tilde{\bm{x}}_{i+1}),\;\; 
    \mathrm{s.t.} \;  \bm{x} \in {}^{\mathcal{E},\mathcal{O}}\sigma_{i}^{\bm{x}} \cap {}^{\mathcal{E},\mathcal{O}}\sigma^{\bm{x}}_{i+1}
    \label{eqn:retarget_x_constraints}
\end{align}
$d(\cdot,\cdot)$ is implemented as the $L_2$ norm of the position difference. This formulation encourages the retargeted object state to stay close to the initial guess from demonstration if possible. $\mathcal{G}_i$ is defined as the $\epsilon$-ball around the solution $\bm{x}$.

\texttt{retarget\_q} find $\bm{q}$ given $\bm{x} \in \leftindex^{\mathcal{E},\mathcal{O}}\sigma_{i}^{\bm{x}} \cap \leftindex^{\mathcal{E},\mathcal{O}}\sigma^{\bm{x}}_{i+1},\mathcal{E,O}$, such that $(\bm{x},\bm{q})\in\leftindex^{\mathcal{E},\mathcal{O}}\sigma_{i+1}$. This is summarized by the following IK feasibility problem:
\begin{align}
    \min_{\bm{q}}\; 0, \;\;
    \mathrm{s.t.} \;  \bm{q} \in {}^{\mathcal{E},\mathcal{O}}\sigma_{i+1}^{\bm{q}\mid\bm{x}}.
    \label{eqn:retarget_q_constraints}
\end{align}
The usage of these subroutines is summarized in Algorithm~\ref{alg:compose_policy}. Implementation of the state constraints in Equations~\ref{eqn:retarget_x_constraints} and ~\ref{eqn:retarget_q_constraints} is described in Section ~\ref{subsec:implement_sigma}.

\subsection{Demonstration collection}
\label{subsec:demon_collection}
The key information to be extracted to the demonstration are the object state trajectory $\left\{\bar{\bm{x}}_t\right\}_{t=0,\dots,T}$, the associated contact sequence $\left\{\sigma_t\right\}_{t=0,\dots,T}$, and the contact transitions $\left\{t_i\right\}_{i=1,\dots,N-1}, \bar{\bm{x}}_{t_i}\subset {}^{\mathcal{\bar{E},\bar{O}}}\mathcal{X}_{s}$. Not needing $\bar{\bm{q}}$ explicitly allows great flexibility in how the demo is obtained.
In this paper, we simply have a human manipulate the object. The top row of Fig.~\ref{fig:highlight_retrieval} shows an example of demonstrating \textit{retrieval} on \textit{cracker}. The human also provides the primitive label. Each demo takes fewer than 30 seconds to complete.
\subsection{Policy composition and execution}
\label{subsec:policy_comp}
Algorithm ~\ref{alg:compose_policy} summarizes an $N$-primitive demo is retargeted to the test time $\mathcal{E,O}$ and object initial state $\bm{x}_0$. Each primitive is given a concrete objective using \texttt{retarget\_x}. Upon completion of the current primitive and prior to executing the next primitive, we leverage \texttt{retarget\_q} at each (freestanding) contact switch state to compute the robot-object contact for the next primitive. We assume the availability of an additional subroutine \texttt{move\_robot\_to}, which relocates the robot to a new state without collision. In our experiments, \texttt{move\_robot\_to} is implemented as a joint position controller, and the robot is always returned to a default configuration prior to moving to the next target \textit{q}.


\begin{algorithm}
    \caption{\texttt{compose\_policy}}
    \begin{algorithmic}[1]
    \Require{$\mathcal{E}, \mathcal{O}, \mathcal{G}, \Pi, \Psi,\left\{\bar{\bm{x}}_t,\sigma_t\right\},\{t_i\}, \bm{x}_0$ }
    \State{$\bm{G}\gets\{\}$}
    \State{$\left\{\tilde{\bm{x}}_{i}\right\}\gets$\texttt{remap\_x}$\left(\left\{\bar{\bm{x}}_{t_i}\right\},\bm{x}_0,\mathcal{E,O}\right)$}
    \For{$i = 0,\dots,N-1$}
    \State{$\bm{G}$.append(\texttt{retarget\_x}$(\sigma_{i}, \sigma_{i+1},\mathcal{E,O}, \tilde{\bm{x}}_{i+1})$)}    
    \EndFor
    \State{$\bm{G}$.append($\mathcal{G}$)}
    \State{$\bm{x}\gets\bm{x}_0$}
    \For{$i = 1,\dots,N$}
        \State{$\bm{q}\gets$\texttt{relocate\_q}($\bm{x}, \mathcal{O,E}$)}
        \State{$\bm{s}\triangleq (\bm{x},\bm{q})$}
        \State{\texttt{move\_robot\_to($\bm{q}$)}}
        \While{$\bm{x}\notin \bm{G}[i]$}
            \State{$\bm{s}\gets \leftindex^{\mathcal{E}, \mathcal{O}}f(\left(\bm{s}, \bm{\pi}_{\sigma_i}(\bm{s}_t;\mathcal{E}, \mathcal{O}, \mathcal{G}_i)\right)$}
        \EndWhile
    \EndFor
    \end{algorithmic}
    \label{alg:compose_policy}
\end{algorithm}

\section{System Setup}
\begin{figure}[b]
 \begin{subfigure}[b]{0.23\textwidth}
         \centering
         \includegraphics[width=\textwidth]{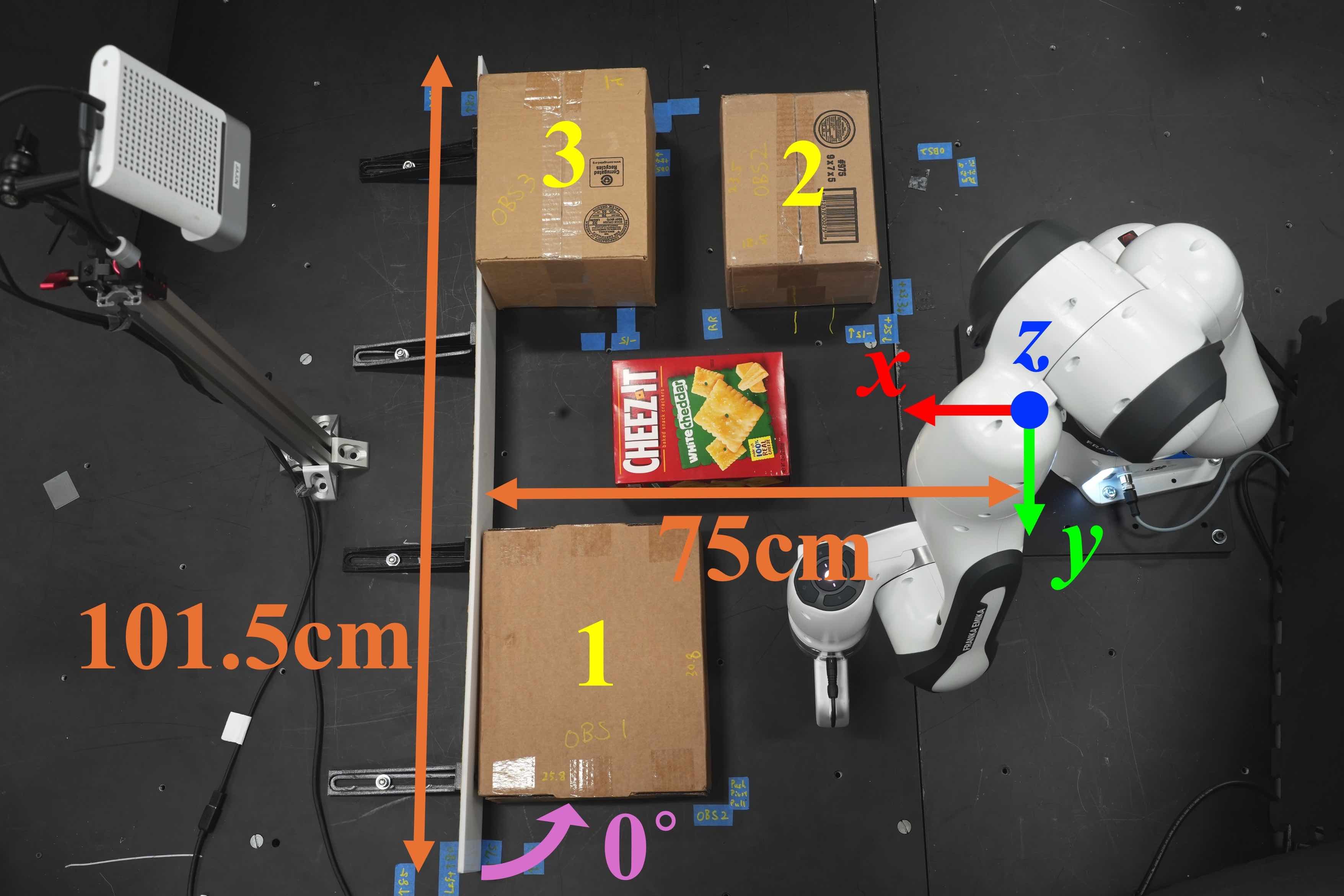}
         \caption{Robot setup}
         \label{subfig:robot}
     \end{subfigure}
     \hfill
     \begin{subfigure}[b]{0.23\textwidth}
         \centering
         \includegraphics[width=\textwidth]{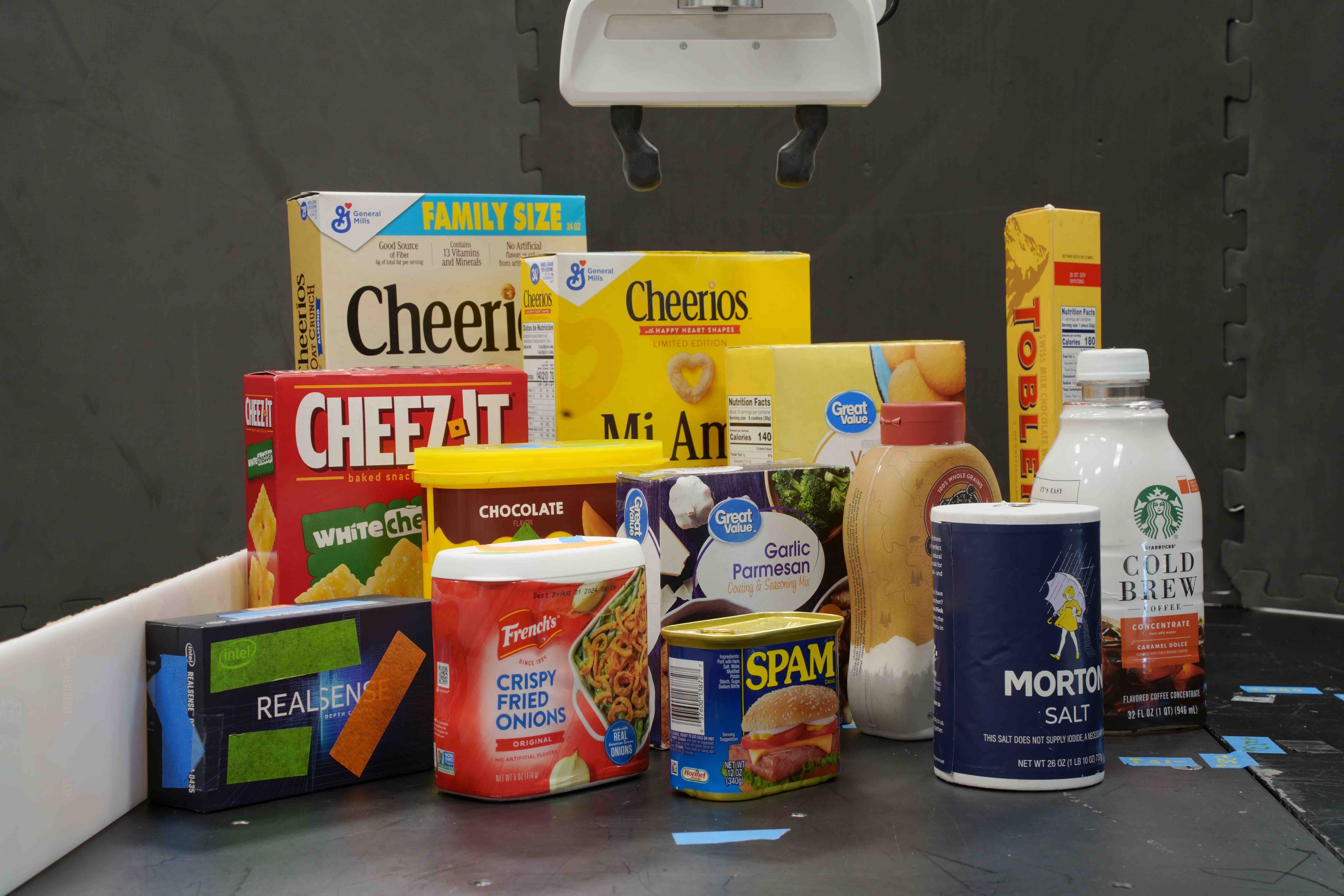}
         \caption{Object set}
         \label{subfig:objects}
     \end{subfigure}
     \hfill
    \caption{\textbf{Hardware setup}. Fig.~\ref{subfig:robot} shows the world frame, the 3 obstacles, and the wall at $(75cm,0\degree)$. Fig.~\ref{subfig:objects} show the 13 objects, from left to right beginning with the frontmost row: camera*, onion*, meat*, salt\dag; cracker, cocoa, seasoning, flapjack, coffee\dag; oat, cereal, wafer, chocolate\dag. *=\textit{short} objects(3). \dag=\textit{impossible} objects(3). The rest are \textit{standard} objects(7).}
    \label{fig:hardware_setup}
\end{figure}

Our hardware setup is shown in Fig.~\ref{fig:hardware_setup}. More details are available on our project website.\\
$\bullet\;\,$\textbf{Robot.} A 2-finger Franka Hand mounted to a 7-DoF Franka Research 3 robot. The gripper has 1 DoF, leading to a 8-DoF action space. Deoxys~\cite{zhu2022viola} is used to control the robot system. We choose the Franka's base frame as the world frame and use it to define the environment configuration.\\
$\bullet$ \textbf{Environment.} A \SI{10}{\cm} tall, \SI{101.5}{\cm} long acrylic ``wall'' is erected at variable distances and orientations from the robot. The wall's pose is expressed with the world $x$ position of its center and its yaw angle about the world $z$, where $0\degree$ is when the wall is parallel to $y$. Up to 3~obstacles may be mounted on the ground in the orientation in Fig.~\ref{subfig:robot}. The dimensions of obstacles $1,2,3$ in cm are $(25.8, 30.8, 7.7), (18.5, 23.5, 14.0), (21.0,25.5,16.3)$. We express the position of an obstacle with the world $(x,y)$ coordinate of its geometric center. \\
$\bullet$ \textbf{Objects.} 7 \textit{standard} objects weighing $120g-400g$ are tested on all tasks. 3 \textit{short} objects are tested on the occluded grasping task. 3 \textit{impossible} objects difficult to manipulate with the gripper are used solely to collect demonstrations in Section~\ref{subsec:retarget_diff_demo}. A textured mesh of each object is captured with Kiri Engine for pose tracking, IK, and training the pushing primitive.\\
$\bullet$ \textbf{Compute.} A desktop computer with Intel i9-13900K CPU and NVIDIA GeForce RTX 4090 GPU is used to run the pose estimation pipeline. A laptop computer with Intel i7-11800H CPU and NVIDIA GeForce RTX 3080 is used to run the primitives.\\
$\bullet$ \textbf{Perception and pose estimation.} A calibrated Microsoft Azure Kinect RGB-D camera is used to capture the scene. Textured meshes of the objects are used to get 6D pose estimations with Megapose~\cite{labbe2022megapose}. 

\section{Experiments}
\label{sec:experiments}
\subsection{Extrinsic manipulation tasks}
\label{subec:ext_manip_tasks}
\begin{figure}[h]
    \begin{subfigure}[b]{0.118\textwidth}
         \centering
         \includegraphics[width=\textwidth]{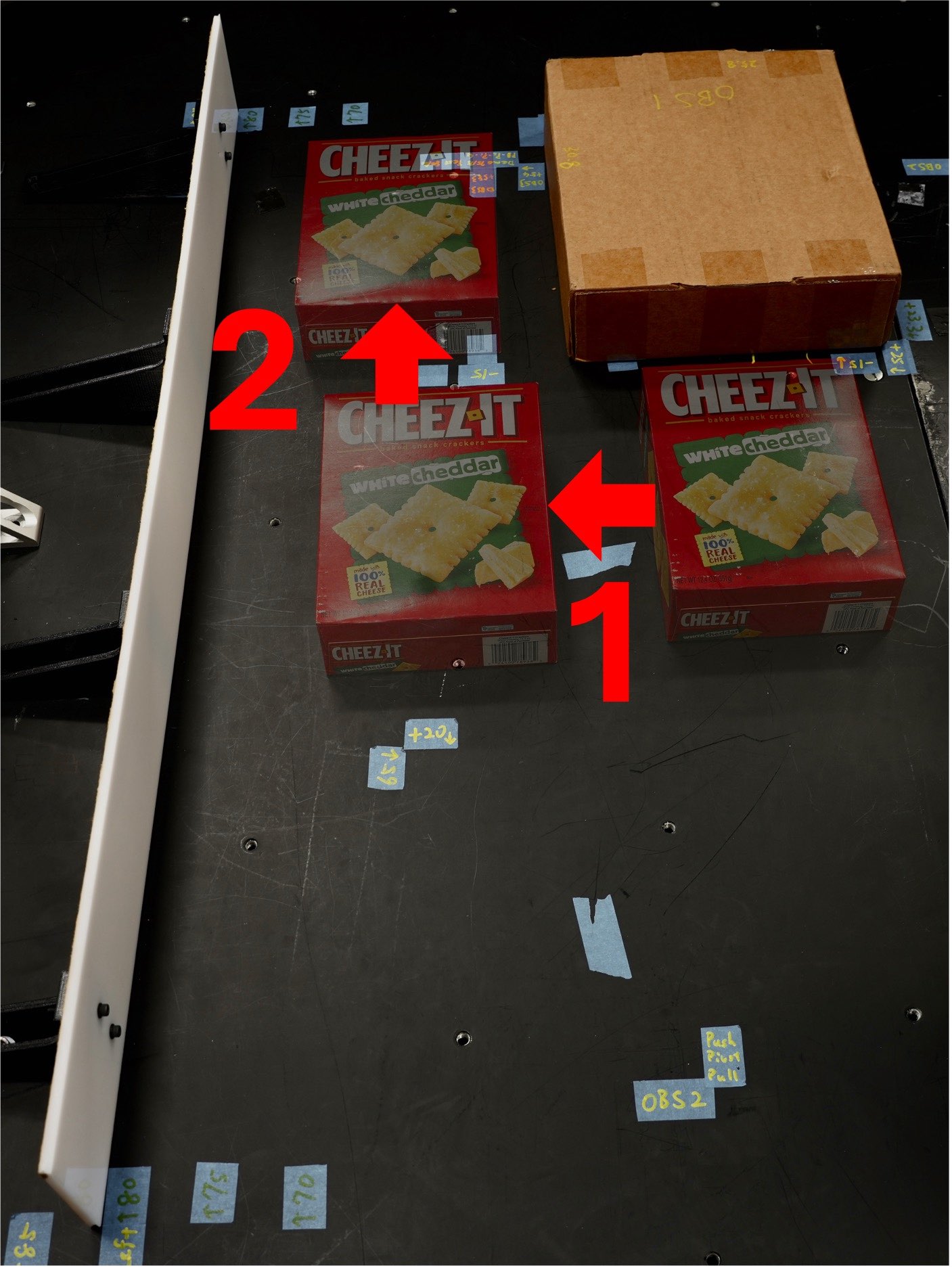}
         \caption{Avoidance}
         \label{subfig:avoidance}
    \end{subfigure}
    \begin{subfigure}[b]{0.118\textwidth}
         \centering
         \includegraphics[width=\textwidth]{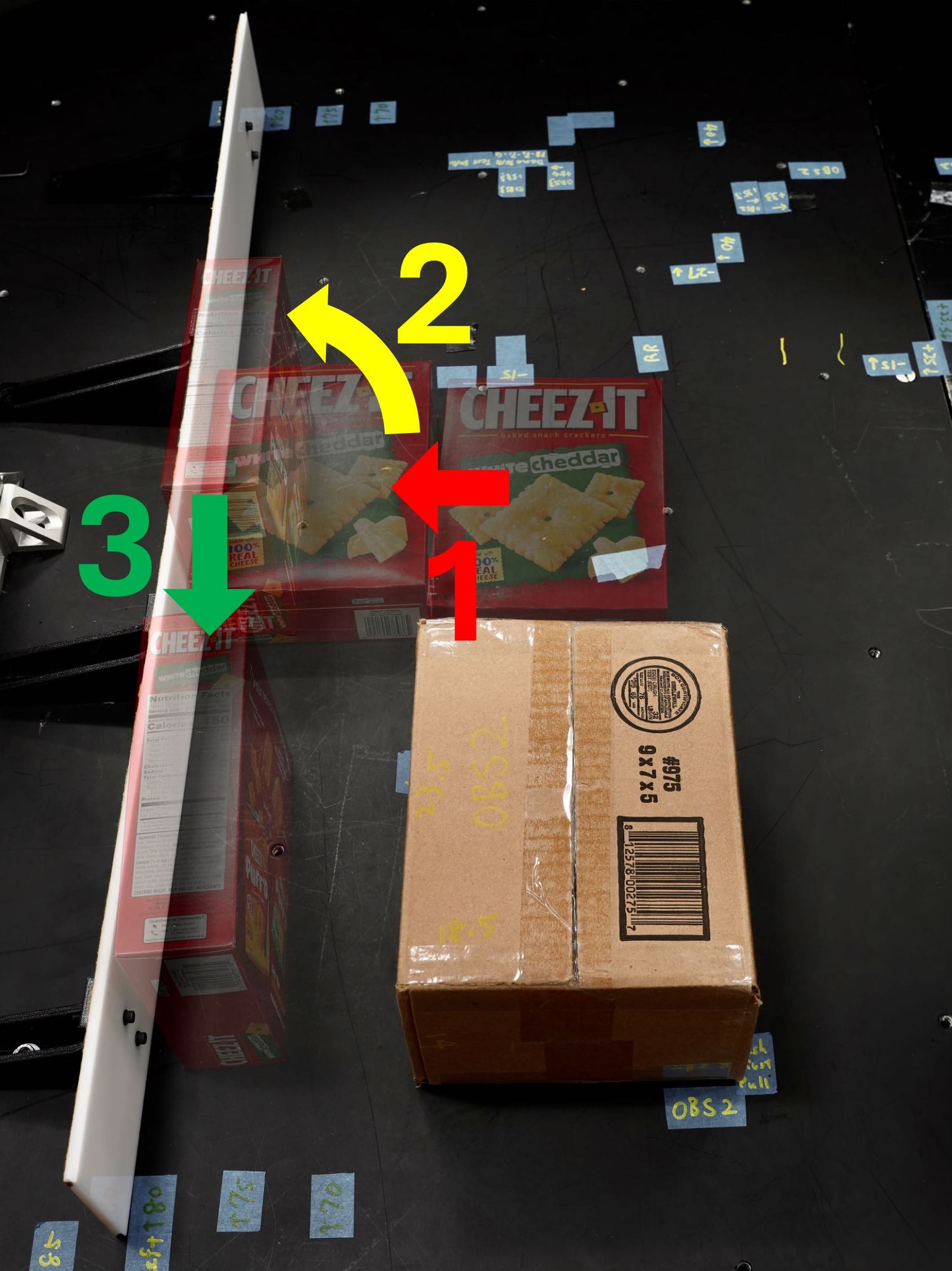}
         \caption{Storage}
         \label{subfig:storage}
     \end{subfigure}
     \begin{subfigure}[b]{0.118\textwidth}
         \centering
         \includegraphics[width=\textwidth]{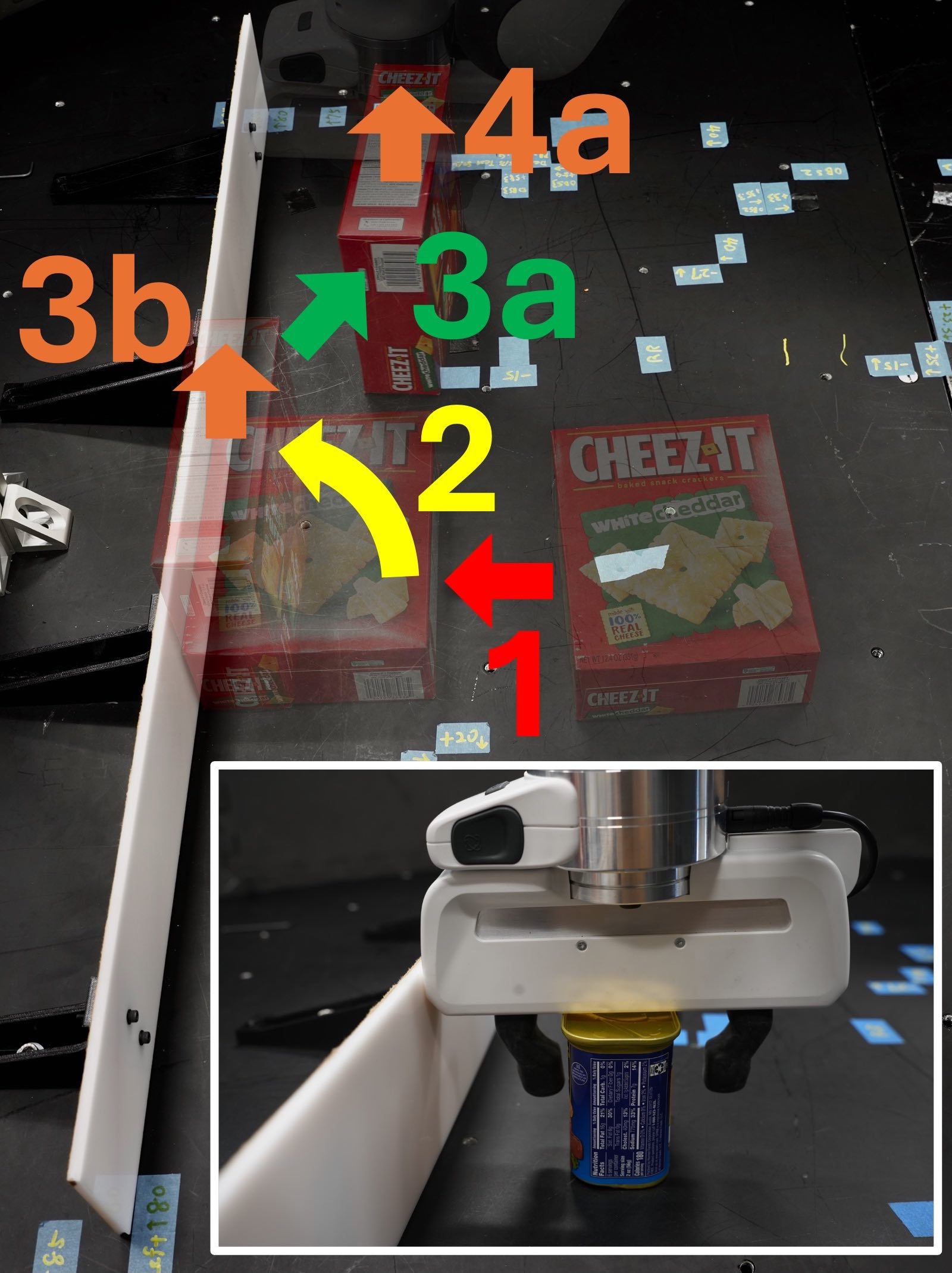}
         \caption{Grasping}
         \label{subfig:grasping}
     \end{subfigure}
     \begin{subfigure}[b]{0.118\textwidth}
         \centering
         \includegraphics[width=\textwidth]{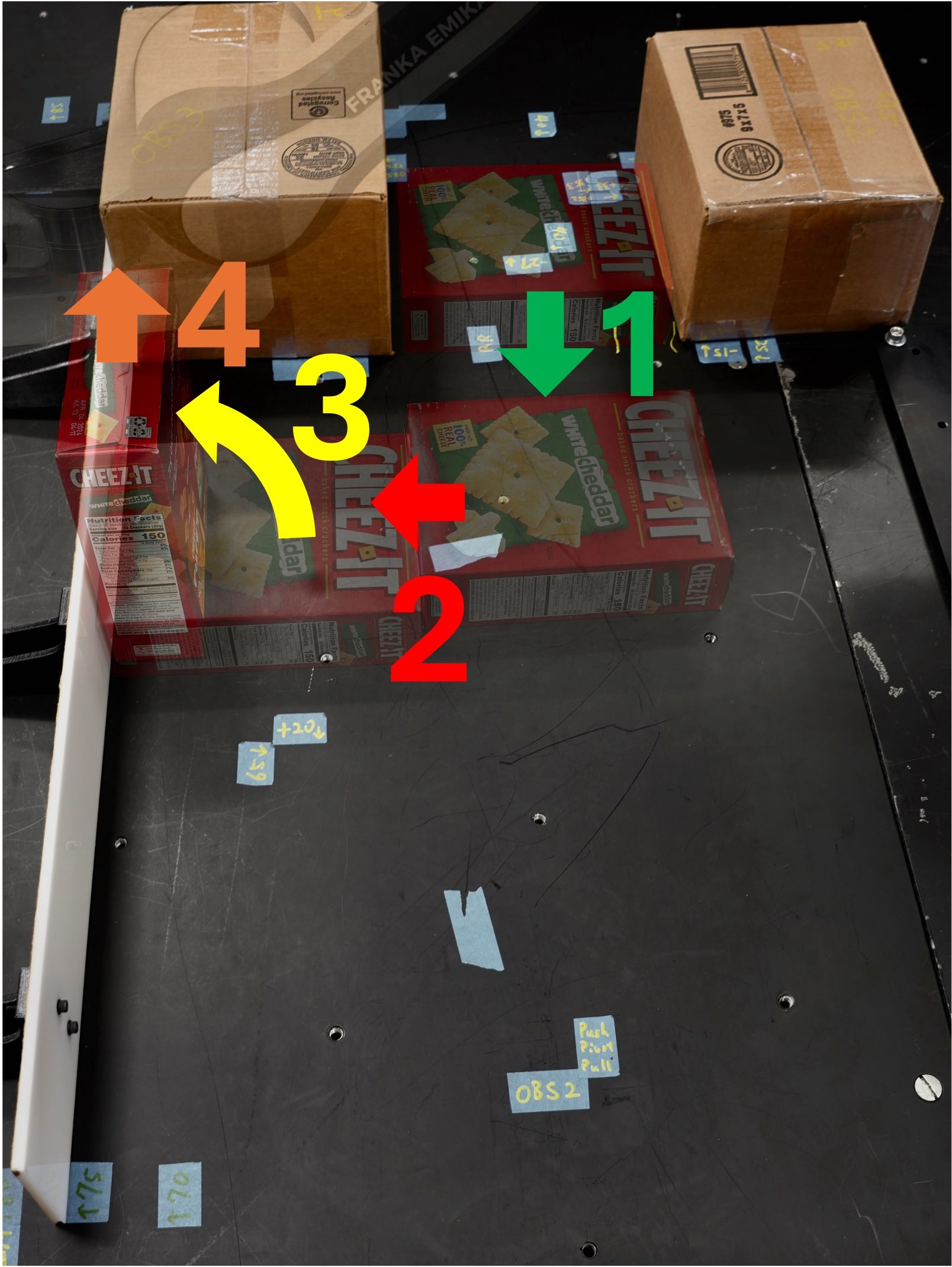}
         \caption{Retrieval}
         \label{subfig:retrieval}
     \end{subfigure}
    \caption{\textbf{Extrinsic manipulation tasks.} The numbers and colors denote the primitive sequence. Push: red. Pull: green. Pivot: yellow. Grasp: orange. An additional ``pull'' is necessary for \textit{short} objects, as the \textit{a} and \textit{b} branches illustrate in Fig.~\ref{subfig:grasping}.
}
    \label{fig:tasks}
\end{figure}

We evaluate our framework on 4 real-world extrinsic manipulation tasks, which are illustrated in Fig.~\ref{fig:tasks}. \\
$\bullet$ \textbf{Obstacle \textit{avoidance}}. \textit{Push} the object forward, switch contact and \textit{push} again to avoid the obstacle.\\
$\bullet$ \textbf{Object \textit{storage}}. \textit{Push} an object toward the wall, \textit{pivot} to align with an opening between the wall and the object, then \textit{pull} it into the opening for storage.\\
$\bullet$ \textbf{Occluded \textit{grasping}}. \textit{Push} the object in an ungraspable pose toward the wall, \textit{pivot} it to expose a graspable edge, and \textit{grasp} it. On \textit{short} objects only, an additional \textit{pull} is performed after \textit{pivot} to create space between the wall and the object for inserting the gripper. To compare with~\cite{zhou2023learning}, a simplified version with no \textit{push} and $\bm{x}_0$ by the wall is also performed on \textit{standard} objects.\\
$\bullet$ \textbf{Object \textit{retrieval}}. \textit{Pull} the object from between two obstacles, \textit{push} toward the wall, \textit{pivot} it to expose a graspable edge, and \textit{grasp} it.\\

Various environments are used for the demonstrations and tests to showcase our method's robustness against environment changes. All demonstrations are collected on \textit{cracker}. Every task is evaluated on the 7 \textit{standard} objects, each with 5 trials. Additionally, \textit{occluded grasping} is evaluated on the 3 \textit{short} with an extra \textit{pull} step.

Numerical results and task parameters are summarized in Table~\ref{tab:task_results} for   \textit{standard} object tasks. \textit{Short} object \textit{grasping} is summarized in Table~\ref{tab:additional_grasping}. Our method achieved an overall success rate of \textbf{80.5\%} (\textbf{81.7\%} for \textit{standard} objects) in Section~\ref{subec:ext_manip_tasks} experiments. Despite not being tailored to \textit{occluded grasping}, we outperformed the equivalent experiments in ~\cite{zhou2023learning}, both when the initial object state is against (\textbf{88.6\%} vs. 78\%) and away from (\textbf{77.1\%} vs. 56\%) the wall.

\begin{figure}[!tb]
    \vspace{6pt}
    \centering
    \includegraphics[width=0.48\textwidth]{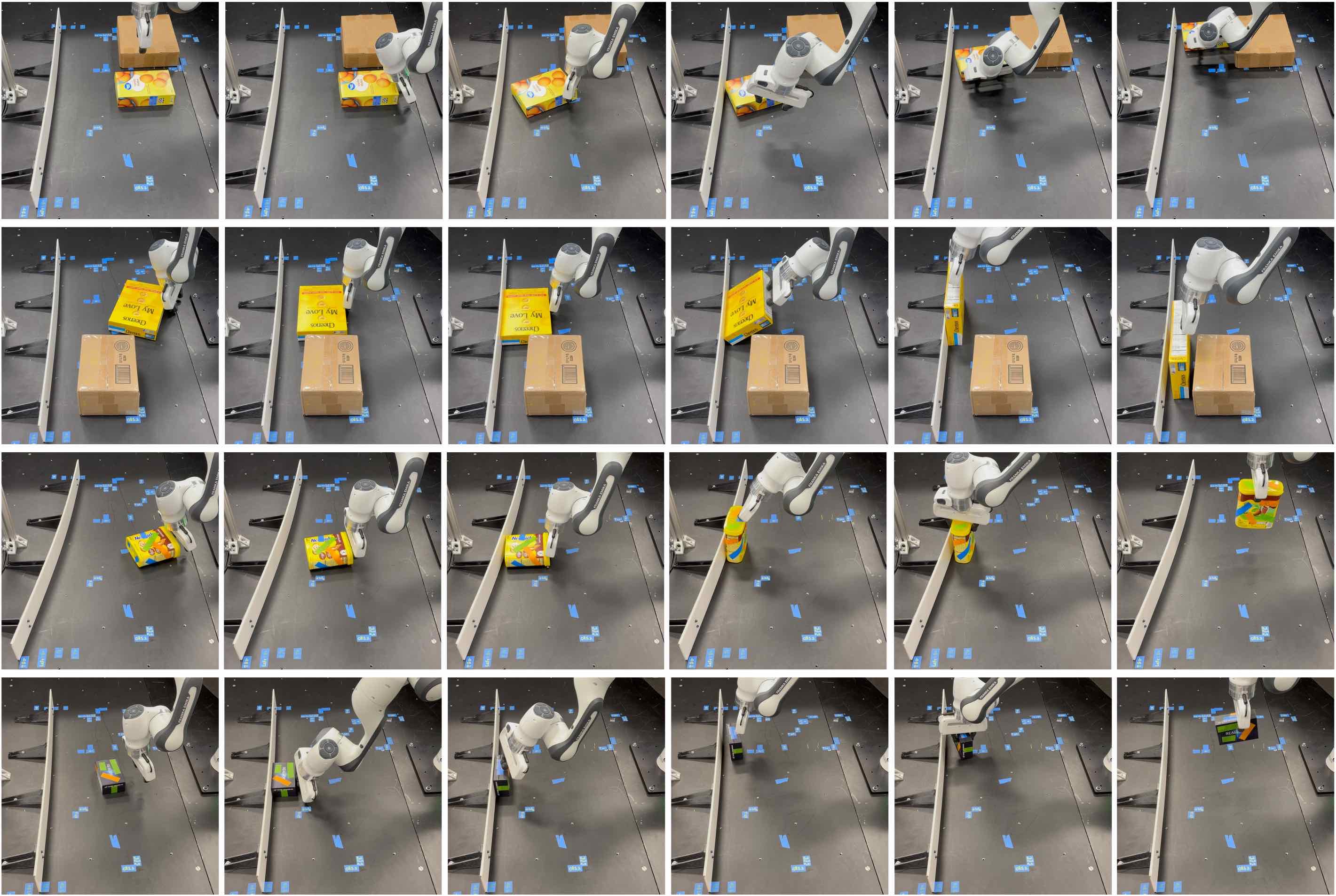}
    \caption{\textbf{Executing extrinsic manipulation tasks}, in temporal order from left to right. From top row to bottom: \textit{avoidance} on \textit{wafer}, \textit{storage} on \textit{cereal}, \textit{grasping} on \textit{cocoa} and \textit{camera}(\textit{short} object). An extra \textit{pull} (5th frame) is necessary to create clearance between the wall and gripper prior to grasping \textit{camera}. Video is available in our supplementary material and on our website. Please refer to Table~\ref{tab:task_results} for detailed task setups and Fig.~\ref{fig:highlight_retrieval} for the \textit{retrieval} task.}
    \label{fig:multi_stage_task}
\end{figure}

\begin{table*}[!t]
\vspace{5pt}
\caption{Summary of experiments on $7$ \textit{standard} objects.}
\label{tab:task_results}
\begin{center}
\begin{tabular}{|c|c|c|c|c|c||c|c|}
\hline
 & Avoidance & Storage & \multicolumn{2}{|c|}{Grasping} & Retrieval & Grasping (Ablation)\\
\hline
Primitives & Push-push & Push-pivot-pull & Pivot-grasp & Push-pivot-grasp & Pull-push-pivot-grasp & Push-pivot-grasp\\
\hline
Demo wall & $80$cm, $0\degree$ & $80$cm, $0\degree$ & $75$cm, $0\degree$ & $75$cm, $0\degree$ & $75$cm, $0\degree$ & $75$cm, $0\degree$\\
\hline
Test wall & $80$cm, $0\degree$ & $75$cm, $0\degree$ & $75$cm, $0\degree$ & $77.5$cm, $-8.5\degree$ & $80$cm, $0\degree$ & $77.5$cm, $-8.5\degree$ \\
\hline
Demo obstacles & $1:(-7.1,14.6)$ & $2:(10.8,31.2)$ & None & None & $2:(-19,49), 3:(23.8,33)$& None\\
\hline
Test obstacles & $1:(-7.1,14.6)$ & $2:(10.8,31.2)$ & None & None & $2:(-19,54), 3:(23.8,35.3)$& None\\
\hline
Cracker & 4/5 & 5/5 & 4/5 & 4/5 & 4/5 & 2/5\\
\hline
Cereal & 4/5 & 5/5 & 5/5 & 5/5 & 5/5 & 3/5\\
\hline
Cocoa & 4/5 & 4/5 & 4/5 & 3/5 & 2/5 & 1/5\\
\hline
Flapjack & 4/5 & 3/5 & 4/5 & 3/5 & 3/5 & 1/5\\
\hline
Oat & 5/5 & 4/5 & 5/5& 3/5 & 5/5 & 2/5\\
\hline
Seasoning & 5/5 & 3/5 & 4/5 & 5/5 & 3/5 & 1/5\\
\hline
Wafer & 4/5 & 5/5 & 5/5 & 4/5 & 4/5 & 3/5\\
\hline
\textbf{Overall} & $\bm{85.7\%}$ & $\bm{82.9\%}$ & $\bm{88.6\%}$ & $\bm{77.1\%}$ &$\bm{74.3\%}$ & $\bm{37.1\%}$\\
\hline
\end{tabular}
\end{center}
\end{table*}

\subsection{Retargeting from different demonstrations}
\label{subsec:retarget_diff_demo}
To show that our method is agnostic to the specific demonstration, we collect demos for \textit{grasping} on \textit{oat} and the 3 \textit{impossible} objects that are unlikely to be graspable by the robot. We then retarget all demos onto \textit{cracker} from 5 different initial poses. We achieved \textbf{100\%} success rate across 20 trials (Table~\ref{tab:additional_grasping}), showing that our method is capable of retargeting from a wide variety of demos.

\begin {table}[H]
\caption{Additional \textit{grasping} experiments.}
\centering
\hspace{-5pt}
\begin{tabular}{ |c|c||c|c|}
    \hline
    \multicolumn{2}{|c||}{\textit{Short} objects} & \multicolumn{2}{c|}{Retargeting to \textit{cracker}}\\
    \hline
    Primitives & Push-pivot-pull-grasp & Primitives & Push-pivot-grasp \\
    \hline
    Demo wall & $80$cm, $0\degree$ & Demo wall & $75$cm, $0\degree$ \\
    \hline
    Test wall & $75$cm, $0\degree$ & Test wall & $77.5$cm, $8.5\degree$ \\
    \hline
    Camera & 3/5 & Chocolate & 5/5 \\
    \hline
    Meat & 4/5 & Coffee & 5/5 \\
    \hline
    Onion & 3/5 & Oat & 5/5\\
    \hline
    - & - & Salt & 5/5 \\
    \hline
    \textbf{Overall} & $\bm{66.7}\%$ & Overall & $\bm{100}\%$\\
    \hline
\end{tabular}
\label{tab:additional_grasping}
\end{table}

\subsection{Ablation and comparison}
\begin{figure}[h]
 \begin{subfigure}[b]{0.33\textwidth}
         \centering
         \includegraphics[width=\textwidth]{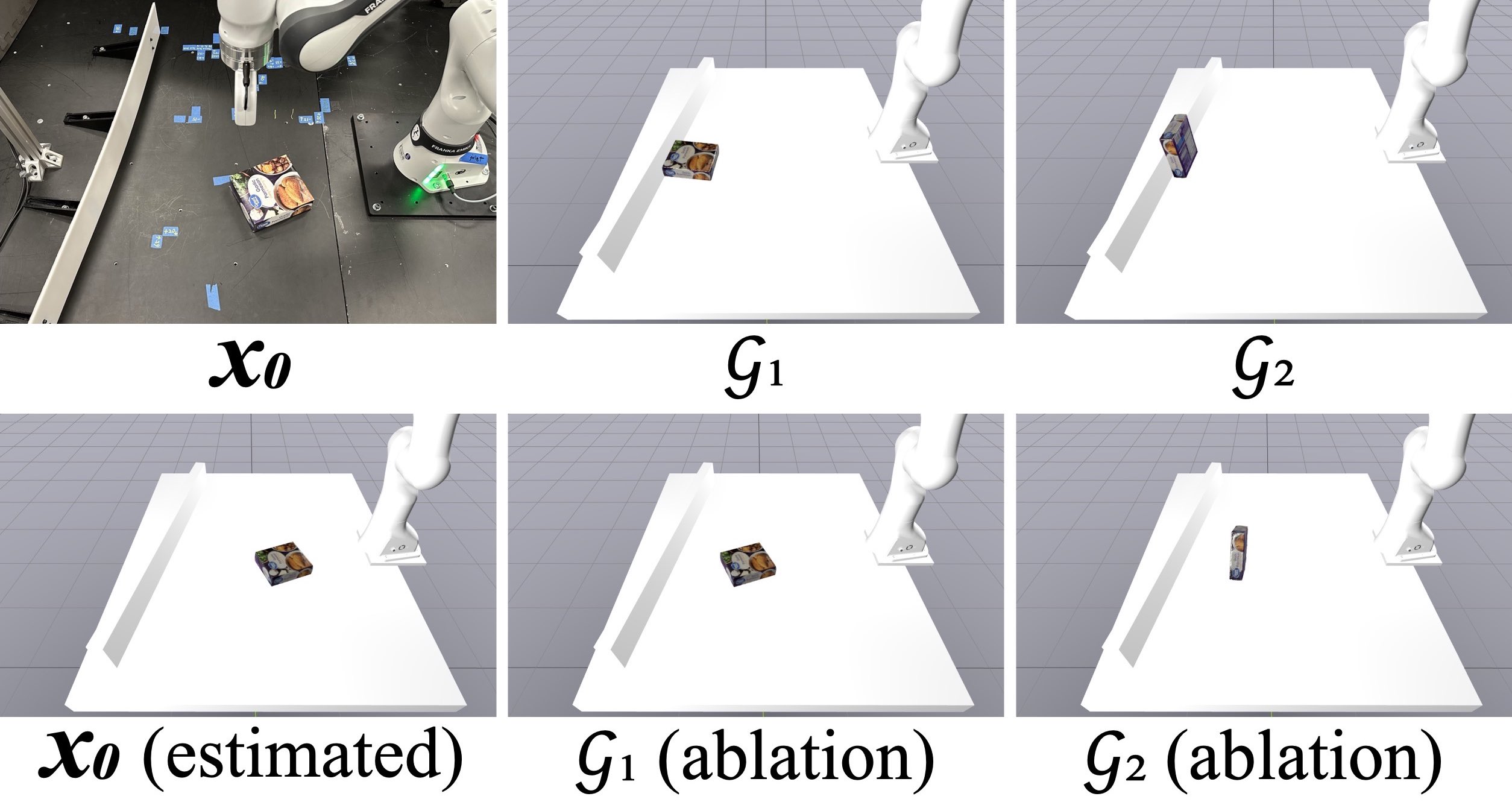}
         \caption{Effect of \texttt{retarget\_x} on $\mathcal{G}_i$}
         \label{subfig:ik_ablation}
     \end{subfigure}
     \begin{subfigure}[b]{0.139\textwidth}
         \centering
         \includegraphics[width=\textwidth]{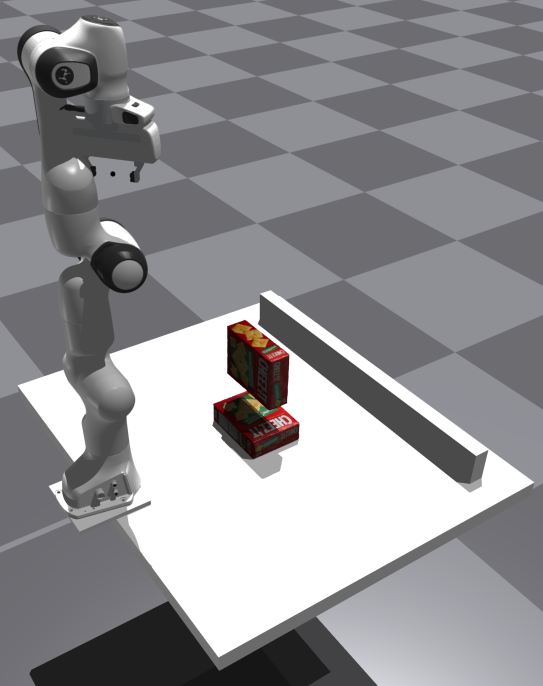}
         \caption{RL scene}
         \label{subfig:rl_scene}
     \end{subfigure}
    \caption{\textbf{Ablation and RL comparison}. Fig.~\ref{subfig:ik_ablation} shows the effects of ablating \texttt{retarget\_x} away on the \textit{grasping} task. The initial state $\bm{x}_0$, \textit{push} and \textit{pivot} intermediate goals $\mathcal{G}_1$, $\mathcal{G}_2$ are shown. $\mathcal{G}_1$, which is just $\tilde{\bm{x}}_1$ as computed by \texttt{remap\_x} without \texttt{retarget\_x}, is too far from the wall for \textit{pivot}.
    Fig.~\ref{subfig:rl_scene} shows the scene, initial (on ground), and goal (floating) poses used to train RL.
    }
    \label{fig:hardware_setup}
\end{figure}

The \textit{occluded grasping} task (\textit{push-pivot-grasp}) is chosen for our ablation study and comparison.\\
$\bullet$ \textbf{Ablation of \texttt{retarget\_x}} shows the merit of contact retargeting. Here, $\tilde{\bm{x}}_i$ are directly set as $\mathcal{G}_i$. This drops the success rate to \textbf{37.1\%} (Table~\ref{tab:task_results}). Without \texttt{retarget\_x}, the intermediate goals $\mathcal{G}_i$ seldom satisfy the contact requirements of the subsequent primitive. This is illustrated in Fig.~\ref{subfig:ik_ablation}.\\
$\bullet$ \textbf{End-to-end reinforcement learning (RL)} serves as a comparison where contact information is not used and the task is treated as one long-horizon task. An RL agent is trained with proximal policy optimization~\cite{schulman2017proximal} to move \textit{cracker} from one initial pose to a goal pose corresponding to a successful \textit{grasping} execution on hardware (Fig.~\ref{subfig:rl_scene}). RL failed to learn meaningful actions using a reward function similar to the one we trained the \textit{push} primitive with. 

\subsection{Failure analysis}
Despite not being a part of our contribution, the perception system and IK solver significantly affect our pipeline's success rate. Poor state estimation and $\mathcal{G}_i$ almost always result in failed execution. A better engineered solution can directly replace our implementation to boost performance. The failure rate of each primitive across experiments in Section~\ref{subec:ext_manip_tasks} are reported below. \textit{Standard} (bold) and \textit{short} objects are reported separately.
\textit{Push}: $\bm{7.5\%}, 0\%$. 
\textit{Pull}: $\bm{4.5\%}, 18.2\%$. 
\textit{Pivot}: $\bm{9.6\%}, 15.4\%$. 
\textit{Grasp}: $\bm{6.6\%}, 10\%$.
We attribute the higher failure rate on \textit{short} objects to perception challenges and the small sizes. Occlusion from the wall is more significant when the object is shorter, and the pose estimation error may be larger. Furthermore, while action noises in our pipeline are of similar magnitude across objects, they also represent a relatively larger impact when the object is small.

\section{Conclusion and future work}
This work presents a framework for generalizing long-horizon extrinsic manipulation from a single demonstration. Our method retargets the demonstration trajectory to the test scene by enforcing contact constraints with IK at every contact switches. The retargeted trajectory is then tracked with a sequence of short-horizon policies for each contact configuration. Our method achieved an overall success rate of 81.7\% on real-world objects over 4 challenging long-horizon extrinsic manipulation tasks. Additional experiments show that contact retargeting is crucial to successfully retargeting such long-horizon plans, and a wide range of demonstration can be successfully retargeted with our pipeline.
Future directions of this work include admitting language-based or simulation-based demonstrations, and generalizing the contact retargeting formulation to remove Assumption~\ref{assump:fs_contact_solution}.









\printbibliography



\end{document}